\documentclass[11pt,a4paper]{article}
\usepackage[nohyperref]{acl2019}
\usepackage{times}
\usepackage{latexsym}
\usepackage{graphicx}
\usepackage{multirow}
\usepackage{booktabs}
\usepackage{amsmath}

\usepackage{url}

\aclfinalcopy 

\title{Incorporating Textual Evidence in Visual Storytelling}

\author{
  Tianyi Li \qquad Sujian Li\\
  MOE Key Lab of Computational Linguistics, School of EECS, Peking University\\
  Peng Cheng Laboratory, Shenzhen, China\\
  {\tt \{litianyi01,lisujian\}@pku.edu.cn}\\}

\date{}

\begin{document}
\maketitle
\begin{abstract}
 Previous work on visual storytelling mainly focused on exploring image sequence as evidence for storytelling and neglected textual evidence for guiding story generation. Motivated by human storytelling process which recalls stories for familiar images, we exploit textual evidence from similar images to help generate coherent and meaningful stories. To pick the images which may provide textual experience, we propose a two-step ranking method based on image object recognition techniques. To utilize textual information, we design an extended Seq2Seq model with two-channel encoder and attention. Experiments on the VIST dataset show that our method outperforms state-of-the-art baseline models without heavy engineering.

\end{abstract}

\section{Introduction}

Multi-image visual storytelling is extended from a long trend of research in image captioning and has attracted considerable attention in recent years. 

To generate the stories, previous work employed a Seq2Seq framework, using image encoder to encode the image sequences and sentence decoder to generate stories from encoded image sequences. 
Most of the researches \cite{1805-05622, 1805-10973, 1806-00738, 1804-09160, 1805-08191, 1708-02977} focused on improving the decoder, and took simple concatenation or an LSTM as encoder. 
With such design, only images are utilized as input in generating the stories.



However, through our observations, the images alone are inadequate for visual storytelling. Storytelling is creative and diversified, so background knowledge is often required to convert a few images to a complete story. However, extracting such background knowledge is very difficult, especially with limited data.

To alleviate such drawback, it is important to take previous experience of story-writing into account. Imagining when a person starts to tell stories from images, he/she may not understand the implications in those images and fail to write a proper story. However, if he/she had heard others telling stories, he/she may be able to tell a story from the stories of similar image sequences he/she previously heard.
 Motivated by such process, we propose to utilize the large corpus as an inventory and improve the visual storytelling model by including stories from similar image sequences in corpus as input to strengthen the encoder design.

On building such models, two major problems need to be solved: (1) how to measure the relatedness of stories from the image sequence pair; (2) how to incorporate the textual information into the model so as to fully exploit it for storytelling.

To handle the first problem of picking the most relevant stories, we propose a two-step ranking method for their image sequences. We first filter out the 'dissimilar' images with object co-occurrence, and then sort the remaining candidates with feature vectors. 
For the second problem of incorporating textual information, we design an enhanced Seq2Seq model with two-channel encoder, one for visual input and the other for textual input. 


We conduct experiments on the VIST dataset \cite{huang2016visual}, a widely used multi-image visual storytelling dataset. We show that with textual evidence, our model outperforms our baselines and state-of-the-art models. 


\section{Method}

\begin{figure}[t]
\includegraphics[scale=0.167]{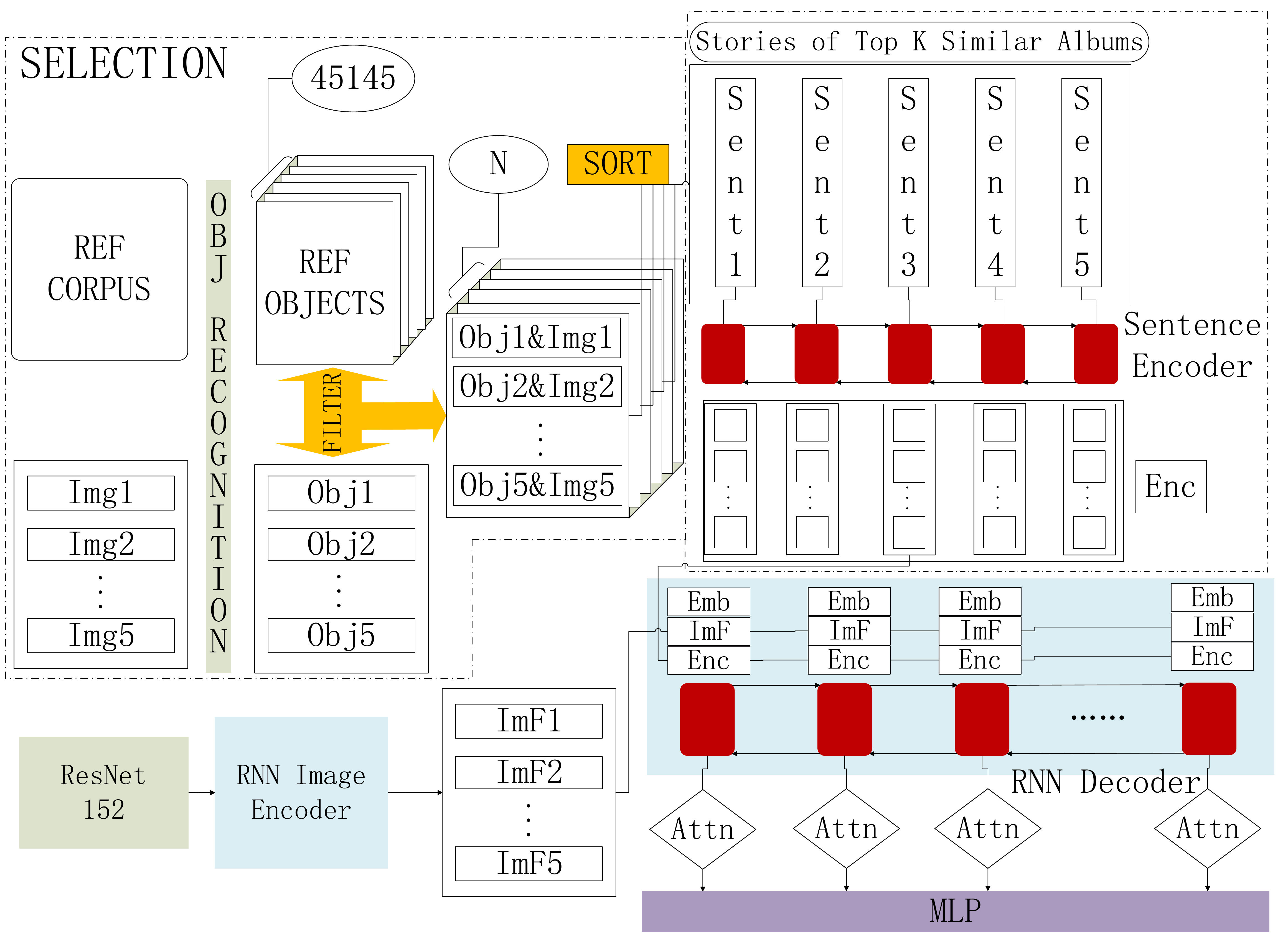}
\caption{Overall architecture of our proposed method. 
}
\label{fig1}
\centering
\end{figure}

Our method is based on the Seq2Seq
framework, composed of a two-channel encoder and a RNN-based decoder.
The whole architecture of our method is shown in Figure ~\ref{fig1}.

In the two-channel encoder, one channel encodes visual evidence from
the image sequence and the other encodes textual evidence from relevant stories. 
In the decoder, we adopt another RNN model to generate stories from the two encoder outputs. To integrate the two types of information, we use Luong attention \shortcite{DBLP:journals/corr/LuongPM15} to dynamically attend to the stories. There are also other modifications, as further explained in ~\ref{2.1}.

To collect the textual evidence for encoder input, we design a selection method described in Section ~\ref{2.2} to get stories from the most similar images.





\subsection{Visual Storytelling Framework}
\label{2.1}
Most previous works on visual storytelling followed the Seq2Seq framework, taking image recognition models
such as ResNet \cite{DBLP:journals/corr/HeZRS15} or Inception \cite{Szegedy_2016_CVPR} 
to extract image features, feeding them into a story-level RNN encoder, bringing encoder output to the sentence-level decoder throughout the generation of the corresponding sentence.

We base our model on this framework with two key modifications: first, we design a text encoder to model the most similar stories which may provide evidence for story generation; second, we adopt the Luong attention \citet{DBLP:journals/corr/LuongPM15} mechanism on the textual side of encoded input to better utilize its information.

\paragraph{Text Encoder}
We use an RNN encoder to model the textual inputs. For each story, we feed its 5 sentences into the RNN one by one, retaining the hidden state across sentences. We take the RNN output of every step through the fully connected layers as encoder output.

\paragraph{Joint Decoder}
Different from previous methods, our decoder depend on both image and text encoder. The incorporation of the two encoders is the key problem. 
Here we adopt two approaches to solve this problem.
First, we use the concatenation of the image encoder output, the embedding of last word and the last hidden states of sentence encoder as the input of the decoder.
Second, we design a Luong attention layer in decoder to attend to sentence encoder outputs.
Formally, the concatenation decoder can be denoted as:
\begin{gather}
s_t^i = DEC(s_{t-1}^i, [emb_{t-1}^i, sent_{len_{sent^i}}^i, img^i])
\end{gather}
and the downstream attention mechanism can be denoted as:
\begin{gather}
weights_t = s_t^i \cdot sent^i \\
C_t = Softmax(weight_t) \cdot sent^i \\
\pi_{\beta}(w_t^i|w_{1:t-1}^i) = softmax(W_c \cdot [C_t, s_t^i] + b_c) 
\end{gather}
where $DEC$ is decoder RNN, $s_t^i$ is RNN output for image $i$ at step $t$, $emb$ is word embedding, $img$ and $sent$ are image and sentence encoder output, $W_c$ and $b_c$ are appended linear matrix and bias.

To be noticed, in our model, both decoder RNN and image encoder are generic and not limited to one particular design. The image encoder can be of arbitrary architecture as long as it generates a vector for each image, and the decoder RNN can also be designed flexibly as long as it takes a vector as input and outputs another vector at each step.

Specifically, we implemented these modifications on two popular systems: GLACNet \shortcite{1805-10973},  the group with best human evaluation scores in Visual Storytelling Challenge NAACL 2018, wwho use residual encoder to generate GLOCAL vectors; XE-ss, a baseline model of \citet{1804-09160}, who proposed to improve performance with reinforcement model (AREL). We call our two models GLAC-TG and XE-TG. 
(see section ~\ref{Sec3.1} for details).

\subsection{Textual Evidence Selection}
\label{2.2}
To provide strong textual evidence for story generation, we aim to select stories which are most similar to the expected story for the given sequence of images. 

With the assumption that similar images usually have similar stories, we take stories of similar images as similar stories. While it's most straightforward to choose the image with the most similar feature vector, it's shown through experiments \ref{select-meteor} that comparing each pair of feature vectors for a large image corpus would be computationally expensive and suffer severely from false positives. Therefore, we propose to employ a two-step filter-and-sort method to pick out the most similar stories.

\subsubsection{Filter}
In the filter step, we use object co-occurrence to discriminate 'roughly similar' image sequences from 'dissimilar' ones. Here we filter by image object information because it conforms with the intuition that images with similar objects describe relevant events. It  is also because object information has been widely used in image captioning as helpful information on images. \cite{1902-09969, 1808-08732, 1804-00887, 1707-07998, 1707-07102, 1805-00314}.

We first get the types and numbers of objects in each image using an object recognition model,
and then we measure image similarity with a categorical criterion and a numerical criterion.
Formally, $O_a$ and $O_b$ are the set of objects present in image $a$ and $b$ respectively, $c_x^k$ is the count of occurrence for object $k$ in image $x$. The categorical criterion concerns the types of common objects, namely $score_{cat} = \frac{|O_a\cap O_b|}{\sqrt{|O_a||O_b|}}$; the numerical criterion concerns the differences in times of occurrence, namely $score_{num} = \frac{|O_a||O_b|}{|\Sigma_{k\in (O_a\cup O_b)}(c_a^k - c_b^k)^2|}$. 
 Additionally, we set similarity scores to $0$ when no objects are recognized in either image.

 As mentioned above, we compare images in sequences. We measure the similarity between the sequences as the average score of its images.
 By filtering on the corpus and keeping only the image sequences scored on the top, we narrow down our candidate sequences to a modest size.

\subsubsection{Sort}
After obtaining a small set of roughly similar image sequences, we  use feature vectors to rank similarity more precisely. Here we experiment on two approaches: a simple cosine similarity measure and a Bi-Linear model with Meteor score as gold annotation inspired by \citet{cao-etal-2018-retrieve}. Empirically we find that Bi-Linear model shows no advantage against cosine similarity. Thus, we simply sort the roughly similar sequences with cosine similarity for downstream models.




\section{Experiments}

\subsection{Experiment Setup}
\label{Sec3.1}
 Our experiment is built on VIST \cite{huang2016visual} dataset, which is organized in 5-image sequences annotated with  5-sentence complete stories. The dataset size is 40098 for train, 4988 for validation and 5050 for test.
 
 In GLAC-TG, we use LSTM RNN model with hidden size  1024, embedding size 256 and learning rate $1\times 10^{-3}$; in XE-TG. We use GRU RNN model with hidden size 512, embedding size 512 and learning rate $4\times 10^{-4}$.
 
 In both models, we use ResNet152 \cite{DBLP:journals/corr/HeZRS15} pre-trained on ImageNet \cite{NIPS2012_4824} as image features, and we use Bi-LSTM and Bidirectional GRU respectively for image encoder. 
 
In both models, we keep the hyper-parameters from their baseline models unmodified. For loss function, we use cross-entropy averaged on the sentence lengths.

 On textual evidence selection, we use all stories and image sequences in train and validation set as reference corpus, and a Fast RCNN \cite{DBLP:journals/corr/HeGDG17, matterport_maskrcnn_2017} model pre-trained on COCO dataset \cite{DBLP:journals/corr/LinMBHPRDZ14} to detect objects from each image. Roughly similar stories are filtered with numerical criterion at 500 candidate size as it shows the best performance.

\subsection{Results}
\begin{table}[!htbp]
  \small
  \centering
    \begin{tabular}{lll}
    \toprule
     {Methods} & {R / C / M} \\
    \midrule
    \citet{huang2016visual} & -\quad\quad-\quad\quad31.4 \\
    \citet{1708-02977} & 29.5\quad7.5\quad34.1 \\
    \citet{1806-00738} & 29.2\quad5.1\quad34.4 \\
    \citet{1805-08191} & 30.8\quad10.7\quad35.2 \\
    \midrule
    GLACNet\shortcite{1805-10973} (re-trained) & 26.3\quad2.2\quad33.0 \\
    \textbf{GLAC-TG-top1(ours)} & \textbf{26.5\quad2.0\quad33.4} \\
    \midrule
    XE-ss\shortcite{1804-09160} & 29.7\quad8.7\quad34.8 \\
    AREL\shortcite{1804-09160} & 29.9\quad8.4\quad35.2 \\
    \textbf{XE-TG-top1(ours)} & \textbf{30.0\quad8.7\quad35.5} \\
    \textbf{XE-TG-top3(ours)}  & 29.6\quad8.3\quad35.4\\
    \textbf{XE-TG-top1-attn(ours)} & \textbf{29.9\quad9.2\quad35.2} \\
    \textbf{XE-TG-top3-attn(ours)} & 29.4\quad9.2\quad35.0 \\
    
    XE-TG-only & 29.1\quad7.7\quad34.8\\
    \bottomrule
    \end{tabular}
    \caption{Performance of our method compared to existing visual storytelling models, R is ROUGE-L, C is CIDEr, M is METEOR (models we re-trained in same setting as original are listed in (re-trained) rows)}
    \label{tab1}
\end{table}

 In Table ~\ref{tab1}, we compare our models with several strong baselines on three automatic evaluation metrics, ROUGE-L, CIDEr and METEOR. In the top block of Table \ref{tab1}, we present 4 previous baselines: 1) a standard Seq2Seq baseline model developed by \citet{huang2016visual}; 2) a hierarchically attentive model designed by \citet{1708-02977}; 3) the Seq2Seq model with sentence-wise separate decoders by \citet{1806-00738}; 4) reinforcement learning with topic guided decoders by \citet{1805-08191}.
 In the middle block, we present the GLACNet model ~\citet{1805-10973} and our improved GLAC-TG model.
 In the bottom block, we present our XE-TG models 
 which are improved based on the XE-ss model in AREL framework \cite{1804-09160}.
 For fair comparison, we evaluate all models with the open source evaluation code\footnote{\url{https://github.com/lichengunc/vist_eval}} \cite{1708-02977}.
 
 Result shows that both our models outperform their corresponding baselines. Even using textual evidence only, our XE-TG-only model shows competitive performance compared to the baselines. Moreover, our XE-TG models using cross entropy loss outperformed state-of-the-art baselines with reinforcement learning techniques \cite{1804-09160, 1805-08191}. By using simple cross entropy loss, our models are also less costly to train, easier to tune and more stable when re-trained. 

\begin{figure}[t]
\centering
\includegraphics[scale=0.192]{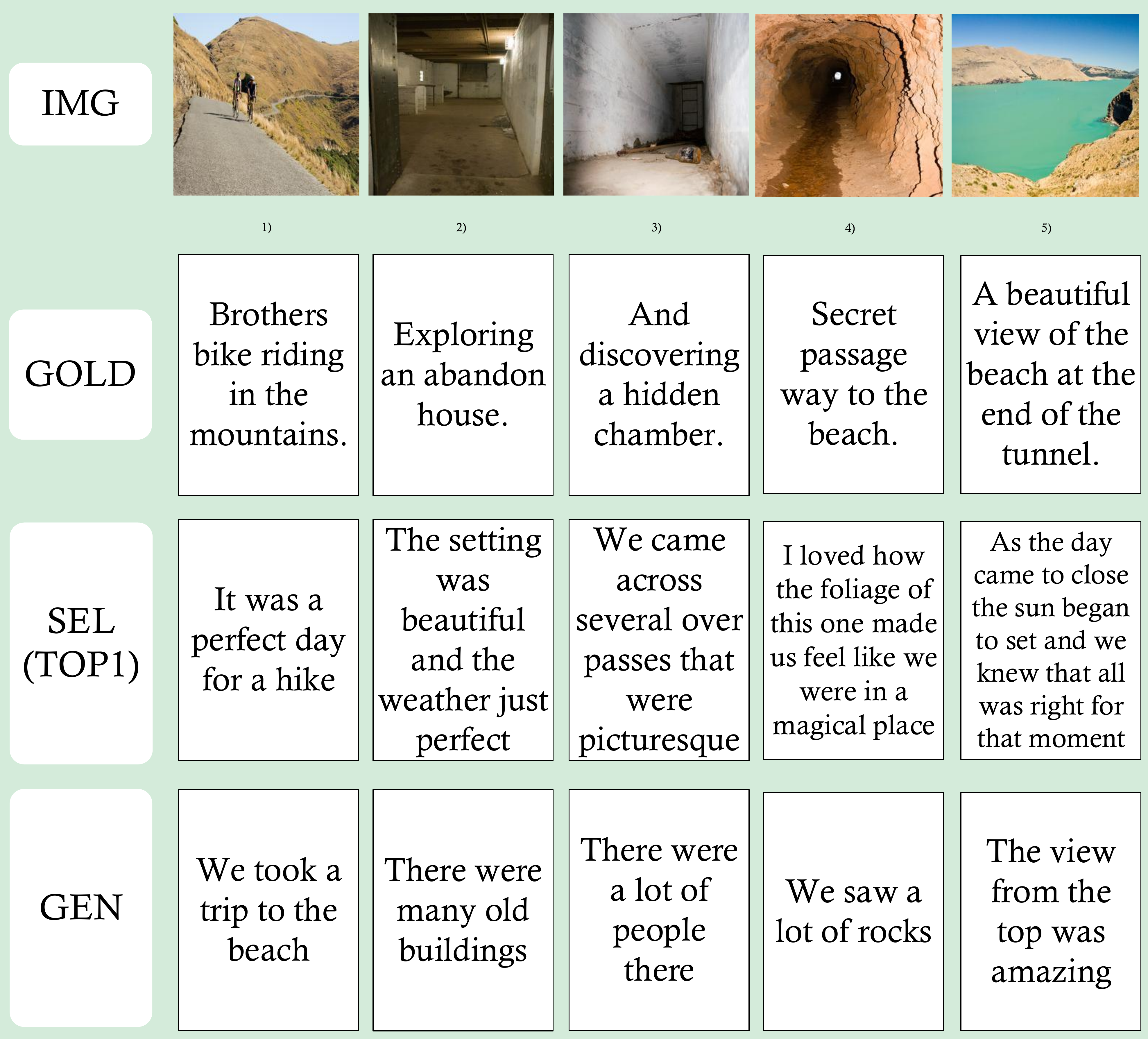}
\caption{An example sequence of visual storytelling.}
\label{fig2}
\centering
\end{figure}

 We conduct a qualitative analysis on XE-TG-top1 model in Figure ~\ref{fig2} as an example. It shows that the selected similar story shares the same topic of wilderness adventure with similar story-flows. The generated story also catches the essence of the image sequence, with basic details closely relevant. It shows that our textual evidence selection method is capable of selecting proper textual evidence, and our storytelling framework is capable of capturing the provided information and telling fluent and coherent stories.
 
 \subsection{Analysis on Textual Evidence Selection}
In this section, we further explore the effectiveness of similar stories.
We experimented on filtering candidate size 50, 100 and 500 with both categorical and numerical criteria, using sorting on the entire reference corpus for comparison and METEOR score as a metric of actual story similarity. In Table ~\ref{select-meteor}, we show that for all methods, the selected stories are significantly more similar to gold stories than randomly selected ones, and stories with higher rankings are generally better than those with lower rankings. Moreover, for both criteria, candidate size poses negligible effect.

On the other hand, neither sorting on full corpus nor sorting by bi-linear model shows competitive results compared to our approach.

\begin{table}[h]
  \centering
  \begin{tabular}{c|c|c|c|c|c|c}
    \toprule
    \multirow{2}{*}{M} &
      \multicolumn{3}{c}{categorical} &
      \multicolumn{3}{c}{numerical} \\
       & {50} & {100} & {500} & {50} & {100} & {500} \\
      \midrule
    1 & 24.8 & 24.8 & 25.0 & 24.9 & 24.7 & 24.5 \\
    2 & 24.9 & 24.8 & 24.7 & 24.4 & 24.5 & 24.6 \\
    3 & 24.6 & 24.5 & 24.6 & 24.6 & 24.6 & 24.5 \\
    4 & 24.5 & 24.9 & 24.8 & 24.5 & 24.5 & 24.3 \\
    5 & 24.8 & 24.6 & 24.6 & 24.5 & 24.5 & 24.5 \\
    rand & \multicolumn{6}{c}{23.8} \\
    full & \multicolumn{6}{c}{23.28 (average on top 5)} \\
    B-L & \multicolumn{6}{c}{23.62 (average on top 5)} \\
    \bottomrule
  \end{tabular}
  \caption{METEOR scores for top 1 to 5 similar stories regarding two criteria, B-L refers to Bi-Linear}
  \label{select-meteor}
\end{table}

\section{Conclusion}

In this paper, we show that textual evidence from similar image sequences contains rich information for visual storytelling, therefore it's capable of boosting storytelling performance. We propose a feasible two-step approach to extract textual evidence from a large corpus. We also design a two-channel encoder to incorporate textual and visual evidence into the Seq2Seq visual storytelling models and achieve state-of-the-art performance without heavy engineering.

\section*{Acknowledgments}
We thank the anonymous reviewers for their helpful comments on this paper. This work was partially supported by National Natural Science Foundation of China (61572049 and 61876009). 

\bibliography{acl2019}
\bibliographystyle{acl_natbib}
\end{document}